\documentclass[10pt,twocolumn,letterpaper]{article}

\usepackage{iccv}
\usepackage{caption}
\usepackage{times}
\usepackage{epsfig}
\usepackage{graphicx}
\usepackage{amsmath}
\usepackage{amssymb}
\usepackage{overpic}
\usepackage{multirow}
\usepackage[table,xcdraw]{xcolor}
\usepackage{enumitem}
\usepackage{soul, color, xcolor}

\graphicspath{{figures/images/}}

\usepackage[pagebackref=true,breaklinks=true,letterpaper=true,colorlinks,bookmarks=false]{hyperref}

\iccvfinalcopy 

\def\iccvPaperID{7256} 

\ificcvfinal\fi

\newcommand\blfootnote[1]{%
	\begingroup
	\renewcommand\thefootnote{}\footnote{#1}
	\addtocounter{footnote}{-1}
	\endgroup
}

\begin{document}

\title{AffordPose: A Large-scale Dataset of Hand-Object Interactions with Affordance-driven Hand Pose}

\author{Juntao Jian$^1$ \qquad Xiuping Liu$^1$ \qquad Manyi Li$^2${$^{\star}$}\qquad Ruizhen Hu$^3$ \qquad Jian Liu$^4${$^{\star}$}\\ 
$^1$Dalian University of Technology \qquad\qquad $^2$Shandong University  \\ $^3$Shenzhen University \qquad\qquad\qquad $^4$Tsinghua University\\
}

\ificcvfinal\thispagestyle{empty}\fi

\twocolumn[{%
\renewcommand\twocolumn[1][]{#1}%
\maketitle
}]

\begin{abstract}
How human interact with objects depends on the functional roles of the target objects, which introduces the problem of affordance-aware hand-object interaction. It requires a large number of human demonstrations for the learning and understanding of plausible and appropriate hand-object interactions. In this work, we present AffordPose, a large-scale dataset of hand-object interactions with affordance-driven hand pose. We first annotate the specific part-level affordance labels for each object, e.g. twist, pull, handle-grasp, etc, instead of the general intents such as use or handover, to indicate the purpose and guide the localization of the hand-object interactions. The fine-grained hand-object interactions reveal the influence of hand-centered affordances on the detailed arrangement of the hand poses, yet also exhibit a certain degree of diversity. We collect a total of 26.7K hand-object interactions, each including the 3D object shape, the part-level affordance label, and the manually adjusted hand poses. The comprehensive data analysis shows the common characteristics and diversity of hand-object interactions per affordance via the parameter statistics and contacting computation. We also conduct experiments on the tasks of hand-object affordance understanding and affordance-oriented hand-object interaction generation, to validate the effectiveness of our dataset in learning the fine-grained hand-object interactions. Project page: \url{https://github.com/GentlesJan/AffordPose}
\end{abstract}

\blfootnote{$^{\star}$ \scriptsize Corresponding Authors: manyili@sdu.edu.cn, jianliu2006@gmail.com}

\section{Introduction}\label{sec:intro}
\begin{figure}[t]
    \begin{center}
    \includegraphics[width=1.0\linewidth]{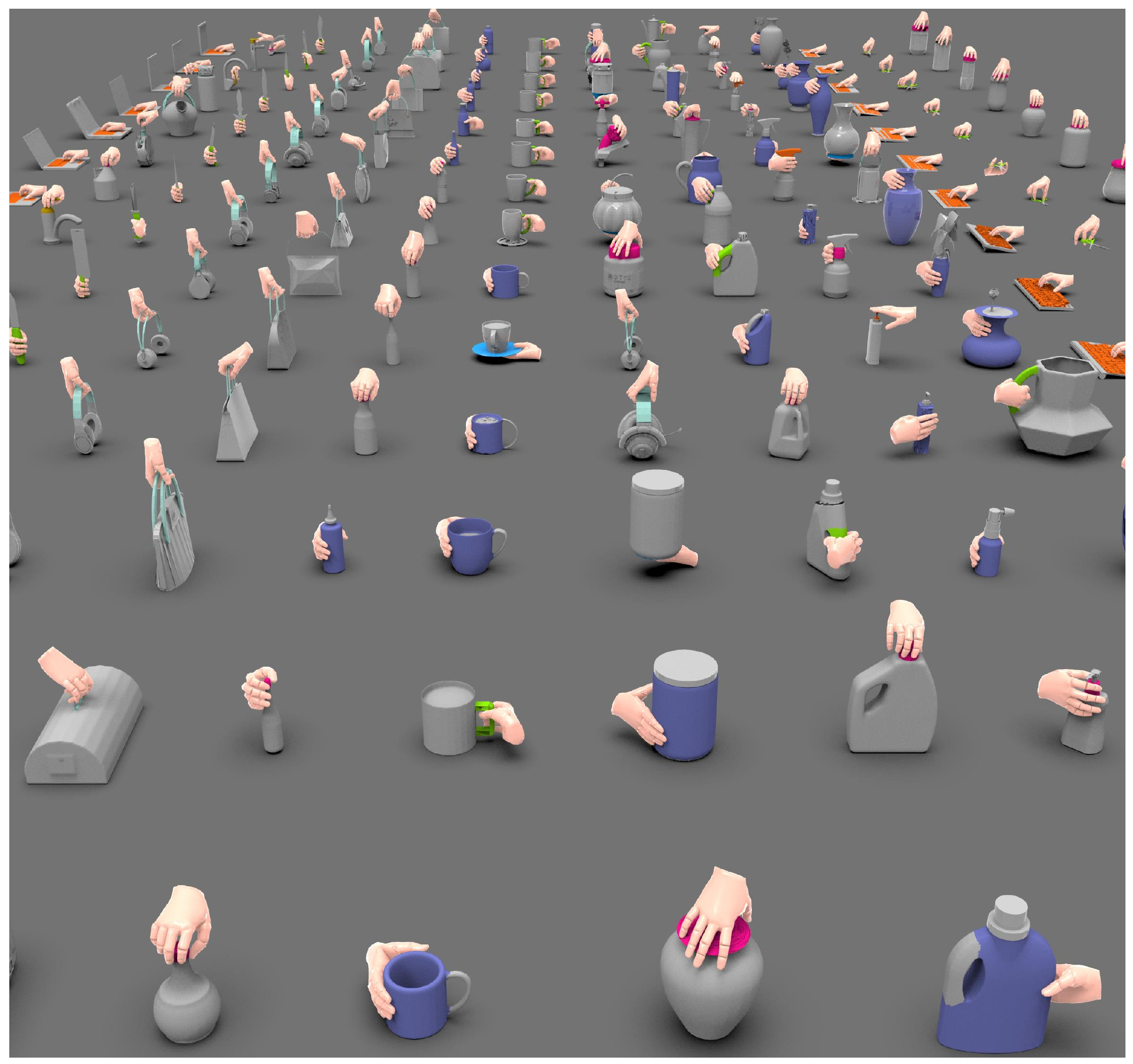}
    \end{center}
    \vspace{-2.5mm}
    \caption{\textbf{A Gallery of AffordPose.} AffordPose is the first large-scale dataset for fine-grained hand-object interactions driven by the specific part-level affordance labeling, which reveals the high correlation between the object affordance and the detailed arrangement of hand poses.}\vspace{-5.5mm}
    \label{fig:teaser}
\end{figure}

One of the long-standing goals of robotics is to imitate all kinds of human-centered interactions, especially hand-object interactions, ranging from general grasping to functional interactions such as unscrewing a cap or even tool usage~\cite{Land1999TheRO,AffordanceSurvey}. Performing appropriate hand-object interaction is a complicated decision-making process. The agents need to understand the functional role of the object, select the contacting location, and perform the specific hand pose to complete the task~\cite{Chu2019LearningAS,Ardon2020AffordancesIR,2016VGaffordance}.

There has been a trend to develop deep learning solutions to predict diverse hand-object interactions. The researchers build hand-object interaction datasets, such as HO-3D~\cite{4_HOannotation}, DexYCB~\cite{5_DexYCB_A_Benchmark_for_Capturing_Hand_Grasping_of_Objects}, Obman~\cite{12_abman}, and train different networks~\cite{16_Xukai,13_Generating_Multi-Fingered_Robotic_Grasps,18_grasp'd,20_Q1} to predict the hand poses for the given objects. However, these works only consider the general grasping task and focus on the stability of the generated hand poses, but overlook the semantic meaning of the hand-object interactions. Recently, many related works collect additional annotations, e.g. contact maps~\cite{8_contactgrasp, 9_ContactPose, Contact2Grasp}, grasp type labels~\cite{11_GanHand}, and intent labels~\cite{3_GRAB:2020, 2022OakInk}, to learn how human use different objects with appropriate hand-object interactions, not only to hold the objects tightly but also for the convenient usage being consistent with human habits.

Object usage involves the key characteristics of the interactions, including what interaction to perform, where to interact with the object, and how to perform the interaction to achieve the purpose. As summarized in Table~\ref{tab:compare_table}, if exists, the object usage, i.e. intents, are often of two types in the relevant datasets. One is human objectives, e.g. use, hand-out, receive as in OakInk~\cite{2022OakInk}, which describes the human targets regardless of the object categories and attributes. Although these human objectives illustrate the purpose of hand-object interactions well, these selected human-centered objectives are general goals and take the objects as just geometric shapes rather than functional objects. The other is object-centric affordances, e.g. pour juice in FPHAB dataset~\cite{1_FPHAB}, which specifies the detailed type of interactions, but with a limited generalization ability. 

In this paper, we take a step further to study the hand-object interactions driven by part-level affordances, which provide fine-grained localizations and are generalizable among object categories. We first collect the specific part-level affordances on the objects, i.e. the hand-centered labels such as twist, pull, handle-grasp, and the corresponding parts, instead of the general labels such as use or handover, then manually adapt the hand poses to complete the interaction tasks corresponding to these affordances. As shown in our dataset, the part-level affordances correspond to some common characteristics of the hand-object interactions even with different object categories, yet allows an extent of hand pose diversity, thus improving the understanding and prediction of hand-object interactions. 

Our AffordPose is a large-scale dataset of fine-grained hand-object interactions with affordance-driven hand poses. The dataset collects 26.7K manually annotated interactions, each including the 3D object shape, the part-level affordance label, and the parameters of the detailed hand configuration. Moreover, our dataset supports the interactions for different affordances on the same object, exhibiting the distinctiveness of the hand poses w.r.t. the corresponding part-level affordance. 

\begin{table}[th]
\centering
\caption{Comparison of AffordPose dataset with the existing hand-object interaction datasets.}
\setlength{\tabcolsep}{0.3mm}{
\small
\begin{tabular}{|c|c|c|c|c|c|}
\hline
dataset      & mod.  & syn/real & \#obj & \#hand  pose & intent                                      \\
\hline
HO3D         & RGBD & real                                                & 10    & 68                                                     & -                                                \\
DexYCB       & RGBD & real                                                & 20    & 1k                                                     & -                                               \\
YCBAfford    & RGB  & syn                                                 & 68    & 367                                                    & -                                      \\ 
Obman        & RGBD & syn                                                 & 2.7k    & 21k                                                      & -                                           \\
FPHAB        & RGBD & real                                                & 26    & 273                                                    & object affordance  \\
ContactPose  & RGBD & real                                                & 25    & 2.3k                                                   & human objective                                                         \\
GRAB         & Mesh & real                                                & 51    & 1.3k                                                   & human objective                               \\ \hline
OakInk-image & RGBD & \multirow{2}{*}{real}                               & 100   & 1k                                                     & \multirow{2}{*}{human objective}  \\ \cline{1-2} \cline{4-5}
OakInk-shape & Mesh &                                                     & 1.7k  & 49k                                                    &                                                 \\ \hline
Ours         & Mesh & syn                                                 & 641   & 26k                                                    & part affordance                         \\ \hline
\end{tabular}}
\label{tab:compare_table}
\vspace{-7pt}
\end{table}

Our contributions are listed as follows:
\begin{itemize}[itemsep=2pt,topsep=0pt,parsep=0pt]
  \item We present the AffordPose dataset, a large-scale dataset of fine-grained hand-object interactions with affordance-driven hand pose.
  \item We provide comprehensive data analysis to understand how affordance affects the detailed arrangement of hand poses to complete the appropriate interaction.
  \item We conduct experiments on two tasks, i.e. hand-object affordance understanding and affordance-oriented hand-object interaction generation, to validate the effectiveness of our dataset in learning the fine-grained hand-object interactions.
\end{itemize}

\section{Related Works}\label{sec:relatedWork}
\begin{figure*}[t!]
\centering
\begin{overpic}[width=1.0\linewidth,tics=10]{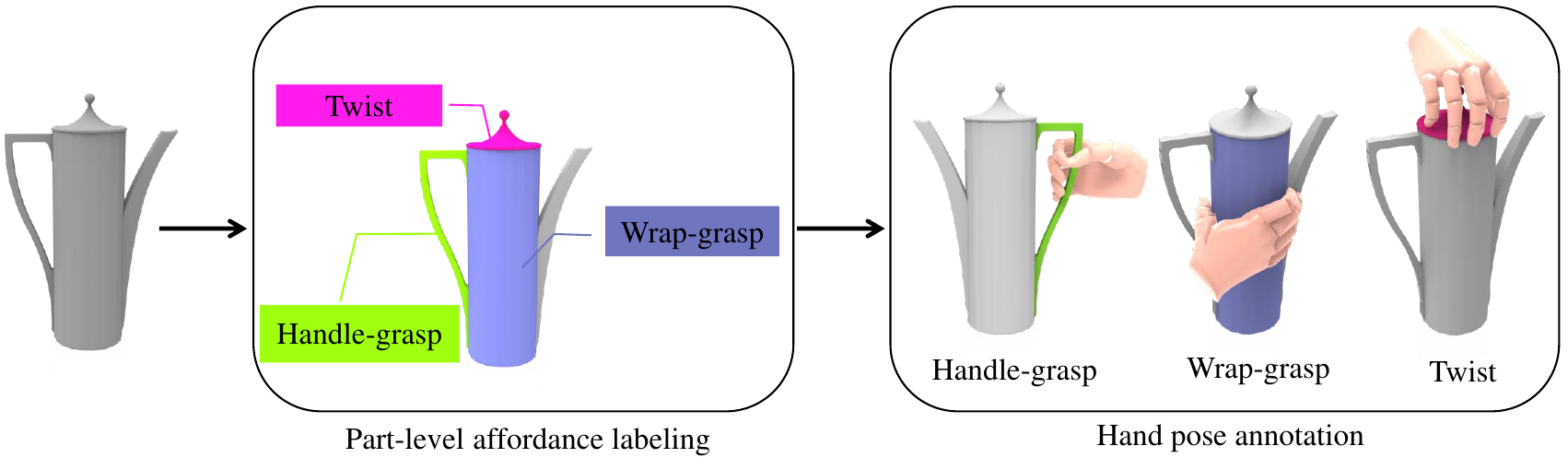}
\end{overpic}
\caption{Dataset construction process for AffordPose. We first annotate the part-level affordance labeling, and then use it as guidance for the volunteers to manually adjust the hand pose annotations.}
\label{fig:pipeline}
\vspace{-7pt}
\end{figure*}


\textbf{Hand-Object Interaction Datasets.}
It's a crucial problem to produce accurate and plausible hand poses to perform hand-object interactions. The researchers have developed different hand pose acquisition, reconstruction, and simulation methods to build large-scale datasets for tasks ranging from hand pose estimation~\cite{Doosti2020HOPENetAG,Liu2021SemiSupervised3H}, and grasp synthesis~\cite{16_Xukai,12_Learning_Dexterous_Grasping}, to hand-object interaction generation~\cite{cao2021reconstructing,11_GanHand}.

Some works focus on hand pose estimation from hand-object interaction observations, e.g. RGB, RGB-D, or video sequential inputs. Therefore, the essential datasets should contain a large amount of accurate hand-object interactions. For example, Bighand2.2~\cite{yuan2017bighand2} collects million-scale hand poses by building a tracking system. Obman~\cite{12_abman} utilizes a grasp optimizer to synthesize the hand poses while ensuring the stability of the grasping. HO-3D~\cite{4_HOannotation} and DexYCB~\cite{5_DexYCB_A_Benchmark_for_Capturing_Hand_Grasping_of_Objects} build the motion capture systems to collect the sequential frames with one or more RGB-D cameras and solves for the 3D hand and object poses to build their datasets. On the other hand, some related works, e.g. DexteriousGrasping~\cite{13_Human-like} and DexGraspNet~\cite{Wang2022DexGraspNetAL}, synthesize for the robotic dexterous hand poses, to complete the grasping task. These constructed datasets provide a wide range of large-scale hand-object interactions for the learning of hand pose estimation from the input observations. 

However, as pointed out in ContactGrasp~\cite{8_contactgrasp}, the hand-object interactions are not only stable but also functional. In other words, the hand-object interaction datasets should involve more human annotations, rather than being built automatically, to reveal how human use different objects. Some related works, e.g. ContactGrasp~\cite{8_contactgrasp}, ContactPose~\cite{9_ContactPose} and Contact2Grasp~\cite{Contact2Grasp}, require the annotators to specify the contact maps of each object and then optimize the hand poses via simulators. The contact map constrains the functional goal of the interactions, while the simulator optimization ensures the physical feasibility of the hand poses. Alternatively, YCB-Affordance~\cite{11_GanHand} requires the annotators to manually specify the hand position, hand pose, and grasp type of each object, and then transfer the grasps to the YCB scenes~\cite{xiang2018posecnn}. These datasets collect natural and realistic hand poses for different objects, which are necessary for the learning of hand-object interaction generation.

Some recent works investigate hand-object interactions with different intents. GRAB~\cite{3_GRAB:2020} captures the whole-body grasps for different interactions, e.g. eating a banana, drinking from a bowl, etc, which are classified into 4 different intents, i.e. use, pass, lift, and off-hand pass. Similarly, OakInk~\cite{2022OakInk} collects affordance-aware and intent-oriented hand-object interactions. That is, the captured hand-object interactions are performed based on the semantic meaning of objects and the specified intents, including use, hold, lift-up, hand-out and receive. H2O~\cite{ye2021h2o} provides a particular benchmark for the human-human object handover analysis.

In our work, we build the dataset named AffordPose, which contains large-scale hand-object interactions with affordance-driven hand poses. Our collected data, termed as affordance-driven hand-object interactions, are performed with the guidance of part-level affordance labels such as twist, pull, handle-grasp, etc. It is different from the grasp type labels in the YCB-Affordance dataset~\cite{11_GanHand} which only indicates different joint arrangements of the hands, or the intents in OakInk dataset~\cite{2022OakInk} which only indicates the general task purpose regardless of the object categories. Although people may consider the object affordances while performing the interactions, i.e. affordance-aware, the corresponding affordance for each interaction is ambiguous and not explicitly specified. By contrast, our dataset contains fine-grained hand-object interactions equipped with the corresponding part-level affordance labeling, which reveals the influence of hand-centered affordances on the detailed arrangement of the hand poses and allows the affordance-oriented hand-object interaction generation.


\textbf{Object Affordance Datasets.} The affordances of objects are related to their functionality, i.e. what it offers the agents or environments~\cite{Gibson1979TheEA}. Hassanin~\cite{Hassanin2021VisualAA} summarizes the common affordances of man-made objects, including the inherent properties (e.g. pour, contain, display, etc) and hand-centered affordances (e.g. twist, pull, press, etc.). Therefore, many object affordance datasets make their efforts in annotating the affordance labels~\cite{2011AAL,2016CAD120,2016CRD,2016HHOI,2017BingeWatching}, aiming at the affordance learning tasks such as the affordance categorization~\cite{2022TVAL,2017CERTH-SOR3D}, detection~\cite{2015UMD,2016VGaffordance,2017IIT-AFF} and segmentation~\cite{Roy2016AMC,Chu2019LearningAS}.

3D Affordance Net~\cite{20213DAffordanceNet} is the first 3D object affordance dataset that provides the dense affordance labels on the 3D point clouds. The per-point affordance distribution highlights the regions where the interactions occur. PartAfford~\cite{2022partafford} focuses on part-level affordance discovery to build the link between the semantic parts and the affordance labels, such as openable, rollable, etc. These works stimulate a deeper understanding of the objects and their affordance, building the basis for various affordance learning tasks, e.g. task-oriented grasp detection~\cite{chen2022learning}, contact map generation~\cite{li2022learning}.

In the spirit of the high correlation between affordance labeling and various hand-object interaction, we further connect the part-level affordance with the detailed configuration of hand-object interactions. In other words, affordance should also reflect how the interactions, especially the hand-object interactions, are performed, which is validated by our data analysis and experiments. Our dataset provides more fine-grained information for the learning of affordance-oriented hand-object interaction.

\begin{table*}[ht]
\centering
\caption{Statistics of the 3D AffordPose Dataset. \textbf{\#Object} denotes the number of objects, \textbf{\#Afford} denotes the number of  affordance and \textbf{\#Hand} presents the number of hand-object interaction annotations.}
\setlength{\tabcolsep}{0.5mm}{
\begin{tabular}{|c|c|c|c|c|c|c|c|c|c|c|c|c|c|c|}
\hline
\multicolumn{1}{|l|}{\textbf{Statistics}} & \textbf{All} & \textbf{Bag} & \textbf{Bottle} & \textbf{Dispenser} & \textbf{Earphone} & \textbf{Faucet} & \textbf{\begin{tabular}[c]{@{}c@{}}Handle\\ bottle\end{tabular}} & \textbf{Jar} & \textbf{Keyboard} & \textbf{Knife} & \textbf{Laptop} & \textbf{Mug} & \textbf{Pot} & \textbf{Scissors} \\ \hline
\textbf{\#Object}                              & 641          & 53           & 52              & 34                 & 50                & 55              & 32                                                               & 45           & 53                & 57             & 50              & 55           & 48           & 57                \\ \hline
\textbf{\#Afford}                                   & 8            & 2            & 2               & 3                  & 1                 & 2               & 5                                                                & 3            & 1                 & 1              & 1               & 3            & 4            & 1                 \\ \hline
\textbf{\#Hand}                                & 26712        & 1624         & 2884            & 2772               & 1400              & 1540            & 2408                                                             & 2716         & 1484              & 1596           & 1400            & 3052         & 2240         & 1596              \\ \hline
\end{tabular}}
\label{tab:unimanual_table}
\end{table*}
\section{AffordPose Dataset}\label{sec:dataset}


\subsection{Dataset Construction}

We recruit volunteers to independently complete the two stages of the data collection, i.e. part-level affordance annotation and hand pose annotation, as illustrated in Figure~\ref{fig:pipeline}. The collected part-level affordance labeling acts as the guidance for the volunteers to complete the following hand pose annotation, to make sure that our collected hand-object interactions match the specific affordance. 

\textbf{Data Preparation.} We collect 641 objects from 13 categories in PartNet~\cite{2018partnet} and PartNet-Mobility~\cite{2020partnet-mobility} to annotate our hand-object interactions. The objects of each category are scaled into their normal size w.r.t. human hands in order to perform the realistic interactions. To create the dataset, we organized a panel discussion with experts and selected 8 hand-centered affordance labels, i.e. the affordance involving hand-object interactions such as press and twist. For the details of object and affordance selection, please refer to our supplementary material. 


\textbf{Affordance annotation.} We manually annotate part-level functional areas of objects to acquire 3D object affordance annotation. In practice, we present the pre-segmented objects to the volunteers and require them to assign the affordance labels to the object parts. The finest level of the hierarchical semantic segmentation from PartNet~\cite{2018partnet} is shown to perform the annotation. For each part of an object, we ask 5 volunteers to discuss and finally make a consensus on the most related affordance label. Note that we only focus on parts with functionality. The volunteers are allowed to mark the other object parts without grasping or manipulating functionalities as "no affordance".


\textbf{Hand-Object Interaction annotation.} In this stage, we present the 3D objects with each part colored based on the affordance label in the visual interface of GraspIt~\cite{GraspIt!} simulator. The affordance labeling illustrates what and where the hand-object interactions should be. The volunteers need to manually adjust the position and rotation of hand palm, and each of the finger joint angles, to complete the appropriate hand-object interaction of each affordance. During the annotation, the object position and orientation are fixed while one can interactively change the viewpoint and adjust the pose of the hand model. We invited 14 volunteers to annotate an average of 42 interactions for each object model, each with at least 28 hand interactions.

The hand model we use is the widely adopted articulated model named MANO~\cite{ManoHand}. Following the practice in~\cite{12_abman}, the hand model is composed of 16 rigid parts, including 3 phalanges of each finger and 1 hand palm. We use the standard hand model with fixed shape parameter $\beta$, while allowing the modification of the pose parameters. The pose parameters include the palm pose (extrinsic parameters) $p = \left\{\textbf{t},\textbf{q}\right\}$ and the joint configuration (intrinsic parameters) $\theta \in \mathbb{R}^{16}$, where $\textbf{t}$ and $\textbf{q}$ represents the transformation and rotation of the entire hand, $\theta$ is the rotation angle of each joint around its pre-defined axis. During the annotation process, similar to YCBAfford dataset~\cite{11_GanHand}, we run the GraspIt! simulator with force analysis to avoid physically implausible hand poses and penetration.


\subsection{Dataset Statistics}

Our AffordPose dataset records the complete hand-object interaction information, including the 3D object shape, the part-level affordance labels to localize the interaction, and the hand pose parameters to complete the interaction task. As listed in Table~\ref{tab:unimanual_table}, the dataset collects a total of about 26.7K affordance-driven hand-object interactions, involving 641 3D objects from 13 different categories and 8 types of affordance, i.e. handle-grasp, press, lift, pull, twist, warp-grasp, support, lever. Each object has 1 to 5 different affordances and each affordance may appear in several object categories, exhibiting rich variation in the collected hand-object interactions. More statistics are listed in our supplementary material.
\section{Data Analysis}\label{sec:analysis}
\begin{figure}[t!]
\centering
\begin{overpic}[width=1.0\linewidth,tics=10]{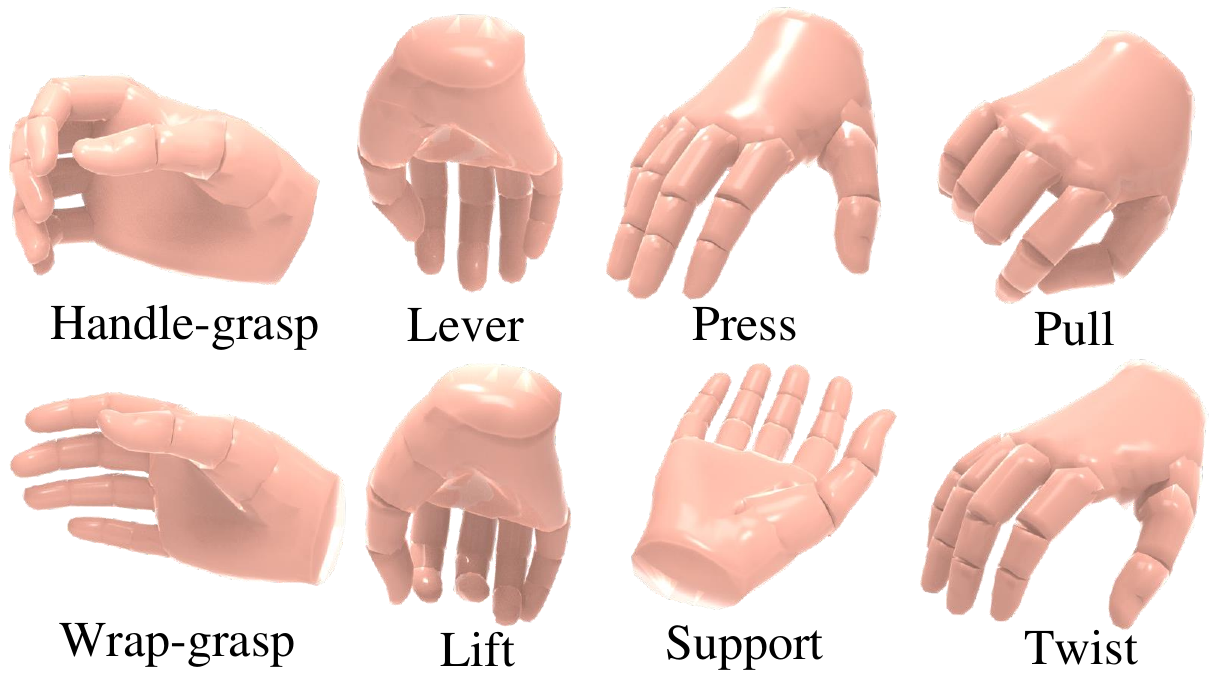}
\end{overpic}
\caption{The representative hand poses for each affordance show their distinctive characteristics.}
\label{fig:representativeHand}
\vspace{-7pt}
\end{figure} 

The hand-object interactions for each affordance exhibit distinctive characteristics and a degree of diversity. We provide the following qualitative and quantitative data analysis to understand how affordance affects the detailed arrangements of hand-object interactions.

\begin{figure}[t!]
\centering
\begin{overpic}[width=1.0\linewidth,tics=10]{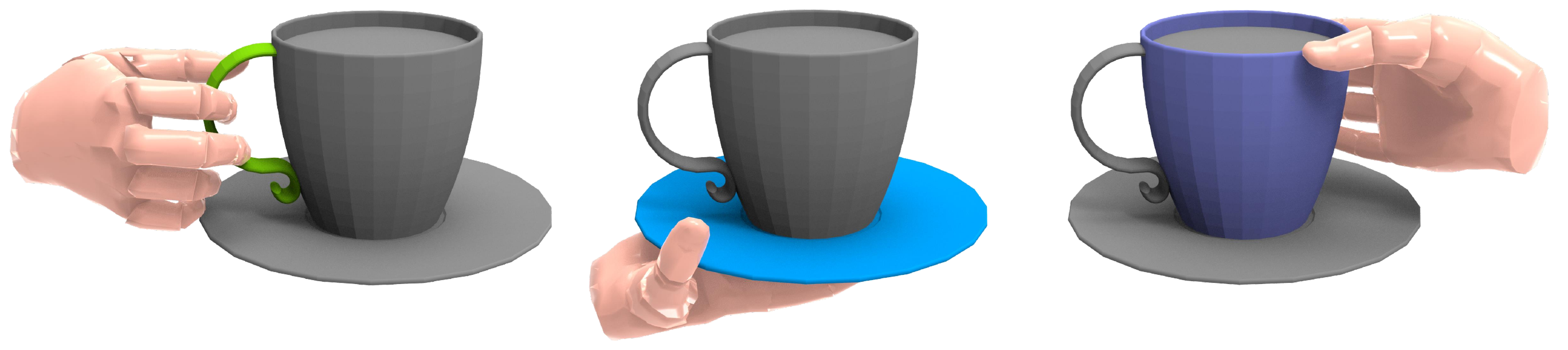}
\end{overpic}
\caption{Different affordances, i.e. handle-grasp, support, wrap-grasp, cause significantly different hand poses to interact with the same object.}
\label{fig:different_affordance_interactions}
\vspace{-7pt}
\end{figure} 

First, the hand poses vary significantly across their different affordances. We show the per-affordance representative hands in Figure~\ref{fig:representativeHand}. The representative hand is defined as the nearest hand model of the mean intrinsic parameters $\theta$ for a specific affordance. Note that we ignore the hand rotation and position (i.e. extrinsic pose parameters) in the computation of the representative hands due to the large variation of the contacting parts from different object categories. Therefore, it only demonstrates the representative joint configurations corresponding to the same affordance label. The representative hands in Figure~\ref{fig:representativeHand} show the distinctive characteristics for different affordances, e.g. the pinching hand for the pull affordance. The distinctivenss of affordance-driven hand poses is also reflected in Figure~\ref{fig:different_affordance_interactions}, where we sample the hand-object interactions on the same mug, but for different affordances such as handle-grasp, support, wrap-grasp. The three hands are completely different in the joint configurations, hand rotations and positions, as well as the contacting parts, w.r.t the corresponding affordances. 

\begin{figure}[t!]
\centering
\begin{overpic}[width=1.0\linewidth,tics=10]{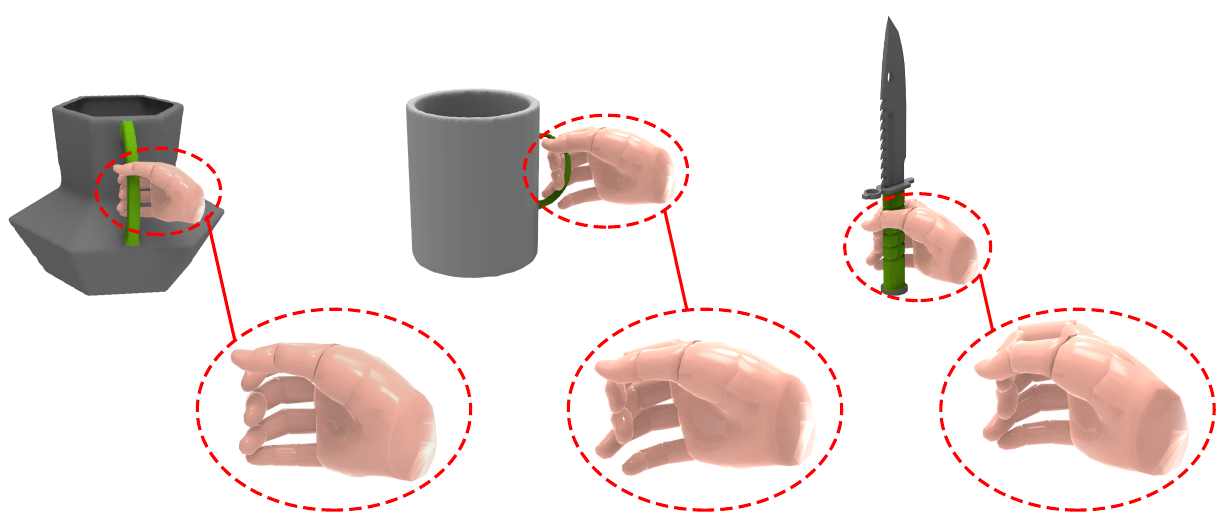}
\end{overpic}
\caption{The hand-object interactions for the handle-grasp affordance on different object categories, i.e. pot, mug, knife, from left to right. The hand poses share similar joint configurations for the same affordance.}
\label{fig:fig_abcdef}
\vspace{-7pt}
\end{figure} 

Second, one affordance often leads to similar joint configurations for the interactions with several different object categories. For example, we show the handle-grasp interactions with a mug, knife, and pot respectively in Figure~\ref{fig:fig_abcdef}. Although the rotation and position of the entire hand (i.e. extrinsic pose parameters) varies when contacting with different parts, the joint configurations of the hands look similar and share some common distinctive characteristics for the handle-grasp affordance.

\begin{figure}[t!]
\centering
\begin{overpic}[width=1.0\linewidth,tics=10]{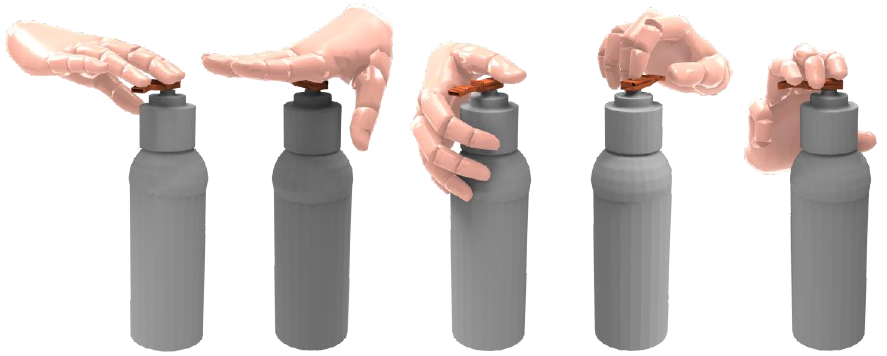}
\end{overpic}
\caption{The hand poses on the same object exhibit some diversities even for the same affordance, e.g. press, due to personal habit factors.}
\label{fig:interactions_with_the_same_object}
\vspace{-7pt}
\end{figure} 

Third, the hand-object interactions of the same affordance also exhibit a degree of diversity, which is mainly caused by different personal habits rather than different object shapes. For example, people may press the lid of the same dispenser object with different contact points on their hands, as shown in Figure~\ref{fig:interactions_with_the_same_object}, resulting in diverse hand poses. However, we can still tell the difference between these hand poses and those for other affordances. 

\begin{figure}[t!]
\centering
\begin{overpic}[width=1.0\linewidth,tics=10]{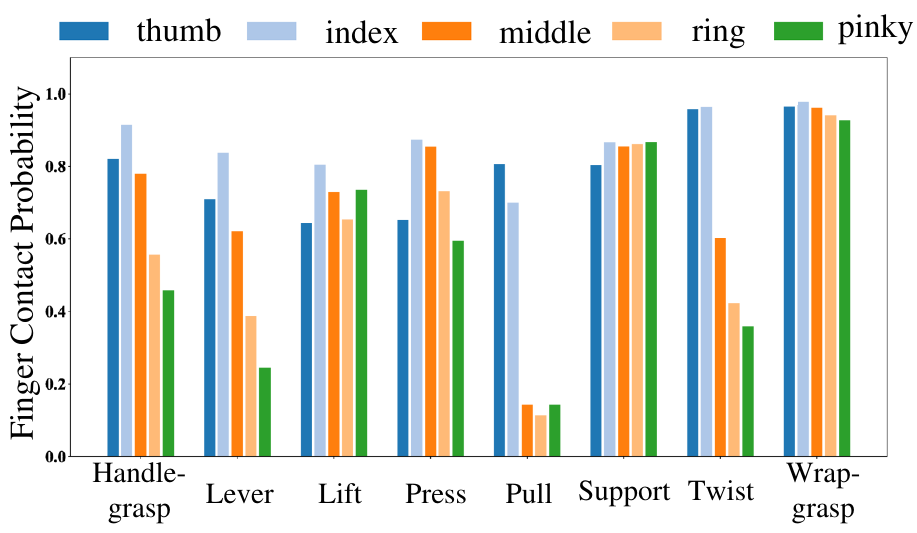}
\end{overpic}
\caption{The probabilities of different fingers contacting the objects for each affordance.}
\label{fig:finger_contact_probability}
\vspace{-7pt}
\end{figure} 
\begin{figure*}[t!]
\centering
\begin{overpic}[width=1.0\linewidth,tics=10]{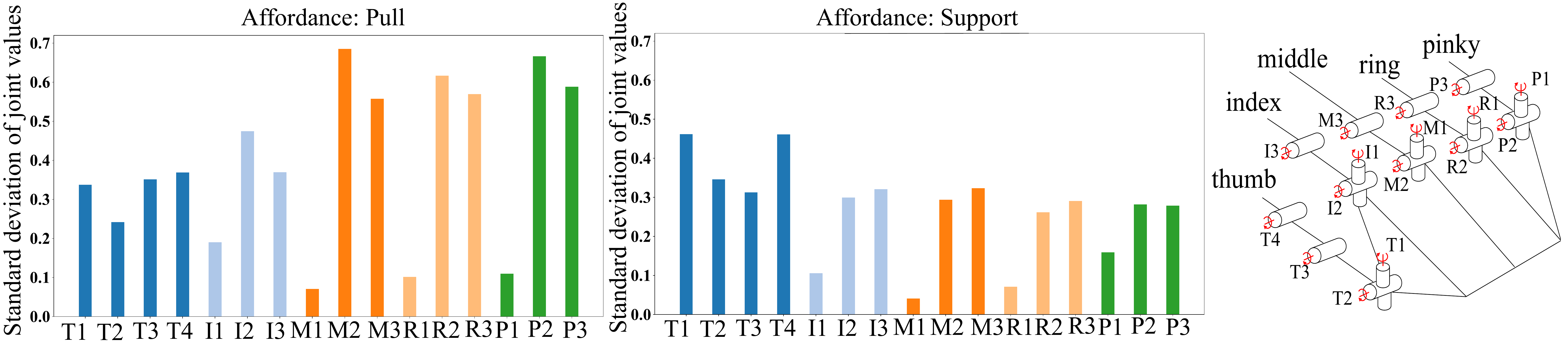}
\end{overpic}
\caption{Left: Standard deviations of joint values for the pull and support affordances. Right: The illustration of 16 DoFs (corresponding to the joint values on the left) to represent the kinematic hand model.}
\label{fig:joint_std}
\vspace{-7pt}
\end{figure*} 

One may be interested in how affordance affects the detailed configuration of the hands, i.e. the arrangement of each finger or joint. We compute the contacting frequency of the fingers for each affordance label (in Figure~\ref{fig:finger_contact_probability}) to understand the importance of the fingers when performing different actions. It shows that the thumb, index finger, and middle finger play a significant role in hand-object interaction for most of the affordances. On the other hand, we further quantify the standard deviation of each hand joint angle, to understand the detailed diversity of each affordance. We specially select the two affordance labels, support and pull, to analyze the per-joint variation in Figure~\ref{fig:joint_std}. We can observe that the DOFs of the root joints (except for the root of the thumb) are often limited for either of the two affordances, i.e. the variations are small for joints I1, M1, R1, P1, but relatively large for I2, M2, R2, P2. Moreover, it's interesting to see how the two quantitative analysis support and complement each other. Taking the pull affordance as an example, Figure~\ref{fig:finger_contact_probability} shows that the thumb and index finger often contact the object, with the other fingers curling inward, while Figure~\ref{fig:joint_std} indicates similar information with lower deviation on the thumb and index fingers and higher deviation on the rest fingers.

\section{Experiments}\label{sec:experiments}
\begin{table*}[ht]
\centering
\caption{The quantitative evaluations of hand-object affordance understanding, including the classification experiments (top two rows) and the part localization experiments (bottom two rows). We report the experiments  with two kinds of hand pose settings as input: the intrinsic pose parameters and all the pose parameters.}
\setlength{\tabcolsep}{2.5mm}{
\begin{tabular}{c|c|cccccccc|c}
\hline
Methods                                                                            &  Inputs         & Handle-grasp & Lever   & Lift    & Press   & Pull    & Support & Twist   & Wrap-grasp & Mean    \\ \hline
\multirow{2}{*}{\begin{tabular}[c]{@{}c@{}}Classification\\ (Accuracy)\%\end{tabular}} & Intrinsic & 92.79      & 99.88 & 98.43 & 98.60 & 99.10 & 91.58 & 90.13 & 93.72    & 94.40 \\ \cline{2-11} 
                                                                                   & All       & 99.50      & 100 & 99.16 & 95.00 & 91.00 & 99.65 & 98.35 & 98.73    & 98.39 \\ \hline
\multirow{2}{*}{\begin{tabular}[c]{@{}c@{}}Localization\\ (IoU)\%\end{tabular}}        & Intrinsic & 95.62      & 96.44 & 97.94 & 94.89 & 77.78 & 88.96 & 90.80 & 97.78    & 95.36 \\ \cline{2-11} 
                                                                                   & All       & 95.05      & 96.99 & 97.90 & 94.42 & 77.78 & 96.59 & 95.17 & 98.91    & 96.29 \\ \hline
\end{tabular}}
\label{tab:exp2_table}
\end{table*}

Our AffordPose dataset contains the 3D objects and the hand poses as demonstrations of hand-object interactions, and the affordance labeling to indicate the fine-grained manipulation purpose. Therefore, it enables two related tasks, i.e. the hand-object affordance understanding and the affordance-oriented hand-object interaction generation. The former aims to understand whether one can infer the corresponding affordance from a rough hand demonstration and a target object to guide the interaction, while the latter is to suggest the corresponding hand-object interaction implementation for a specific affordance-related purpose. We also conducted an RGB-based hand-object interaction understanding and mesh recovery experiments. 


\subsection{Hand-object Affordance Understanding}

Due to the high correlation between the affordances and hand-object interactions, inferring the affordance labeling from a rough hand demonstration and a target object, i.e. hand-object affordance understanding, plays an important role to guide or suggest the users what to perform and where to interact in the following interactions, which is useful in many scenarios such as AR applications. 

We conduct multiple experiments with different input and output settings. The input hand demonstrations are represented as either the intrinsic hand pose parameters (i.e. the joint configurations of the hands) or all the pose parameters including the hand rotation and position as well. The network outputs the object-level affordance label (i.e. affordance classification experiments) or the per-point affordance labels (i.e. part localization experiments). The per-point affordance label is defined as the part index value if the point belongs to the corresponding part to be interacted with, and $0$ otherwise.

\begin{figure}[t!]
\centering
\begin{overpic}[width=1.0\columnwidth,tics=10]{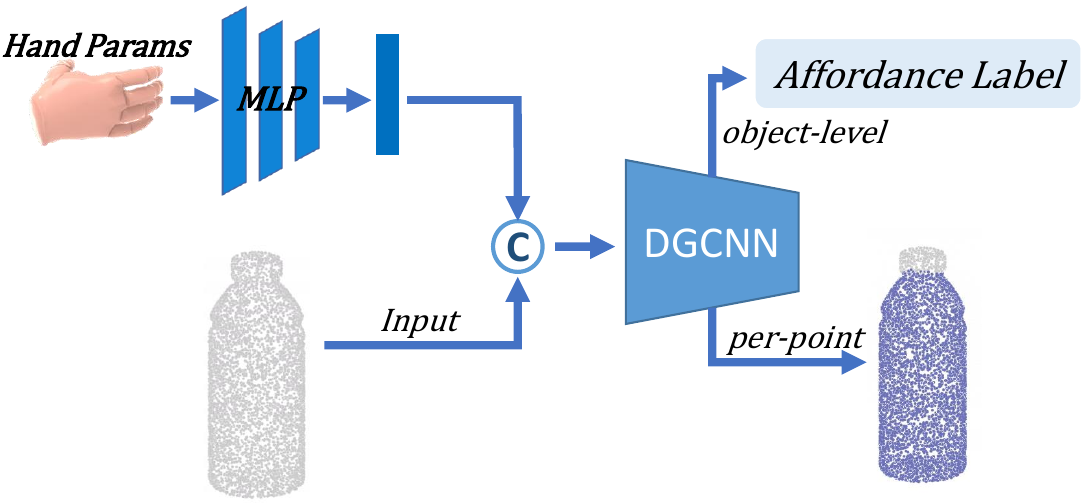}
\end{overpic}
\caption{The hand-object affordance understanding network. It takes a hand pose and a target object as input and predicts the object-level or per-point affordance labels with different branches of DGCNN to guide the interactions.}
\label{fig:5_1_hand_network}
\vspace{-7pt}
\end{figure}

All the experiments are implemented using the network architecture of DGCNN~\cite{6_DGCNN}, which is developed to process the point clouds with its Edge-Conv modules. Specifically, as shown in Figure~\ref{fig:5_1_hand_network}, we encode the hand pose parameters as a 10D vector with a 4-layer MLP network, and concatenate it with the coordinates of each point to form a high-dimensional point cloud $P'\in R^{N\times 13}$, where $N$ is the number of points. We adopt the two corresponding branches of DGCNN~\cite{6_DGCNN}, i.e. the classification branch and the segmentation branch, to output the object-level affordance and the per-point affordance labels respectively. The network is trained on 8-1-1 train-val-test split of the dataset.

\begin{figure*}[t!]
\centering
\begin{overpic}[width=1.0\linewidth,tics=10]{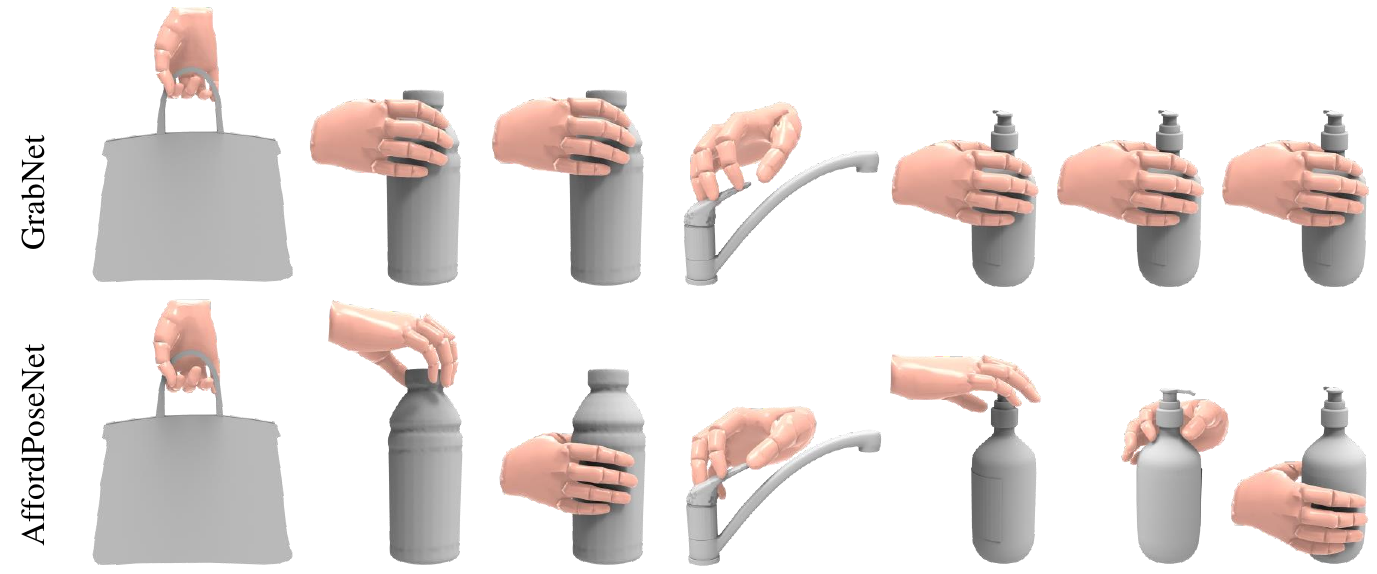}
\put(12,-1.5){Lift}
\put(29,-1.5){Twist}
\put(39,-1.5){Wrap-grasp}
\put(55,-1.5){Lever}
\put(71,-1.5){Press}
\put(82,-1.5){Twist}
\put(92,-1.5){Wrap-grasp}
\end{overpic}
\caption{Qualitative results of hand-object interaction generation experiments. Top: GrabNet baseline with only the object as input. Bottom: AffordPoseNet with both the object and affordance condition as input. For the objects which have several different affordances, GrabNet fails in generating diverse hand-object interactions, while AffordPoseNet is able to generate the appropriate hand poses corresponding to the given affordance.}
\label{fig:exp1}
\vspace{-7pt}
\end{figure*}
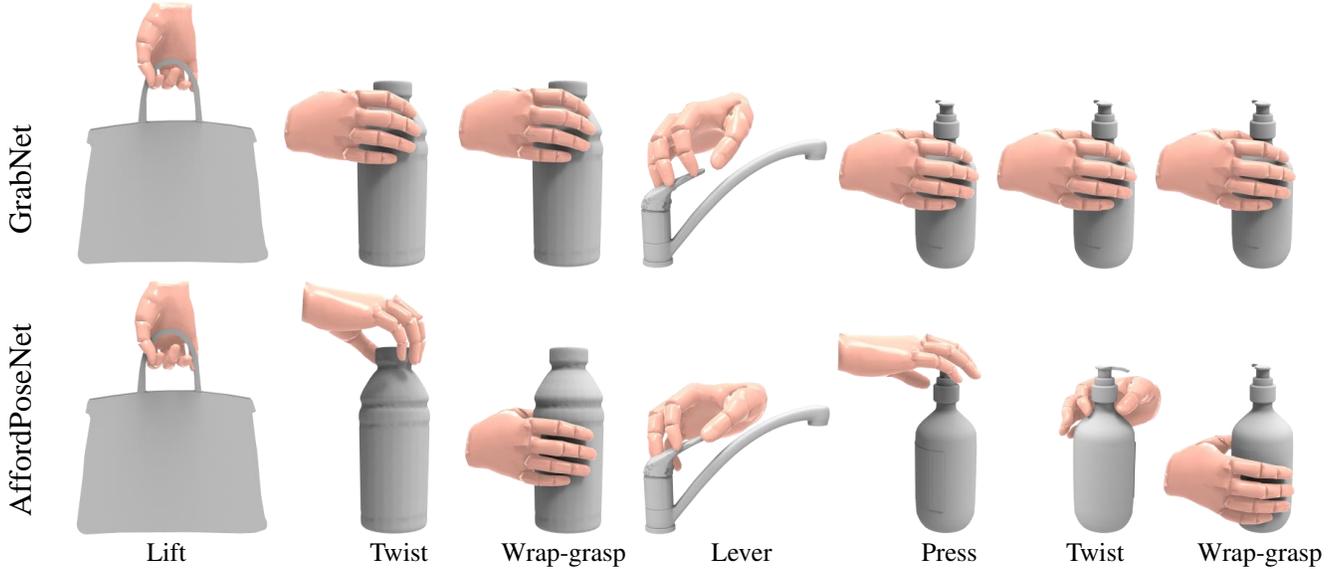 

The quantitative evaluations are reported in Table~\ref{tab:exp2_table}. We compute the affordance accuracy and IoU to measure the performance of the classification (top two rows) and the parts localization (bottom two rows) respectively. The IoU reflects whether the predicted per-point affordance label matches with the region of the corresponding part for the given hand pose. Overall, all the experiments achieve quite high performance, which aligns with our conclusion in the data analysis that there's a high correlation between affordance and hand poses. In addition, the mean affordance labeling from all the parameters consistently outperforms that from the intrinsic parameters only. This implies that although different affordances correspond to certain types of hand joint configurations, it is also affected by the rough rotation and position of the input hand demonstration. 

The per-affordance accuracy of these experiments listed in Table~\ref{tab:exp2_table} offers more detailed evaluations. For example, the pull affordance obtains the worst performance in all the experiments except for the affordance classification from intrinsic parameters. This is because the main characteristics of these hand poses are the joints of the thumbs and index fingers, included in the intrinsic parameters. However, the contacting part of the pull affordance, i.e. zippers of bags, is relatively small, which affects the performance of the part localizing experiments. Additionally, both the classification and part localization performances of the support and twist affordances are small when taking only the intrinsic parameters as input, but are largely improved when all the parameters are used. The reason might be that the two affordances usually share similar joint configurations, which are represented by the intrinsic parameters, but different hand rotations and positions, which are recorded in the extrinsic parameters of the hand poses.

\begin{figure}[t!]
\centering
\begin{overpic}[width=1.0\columnwidth,tics=10]{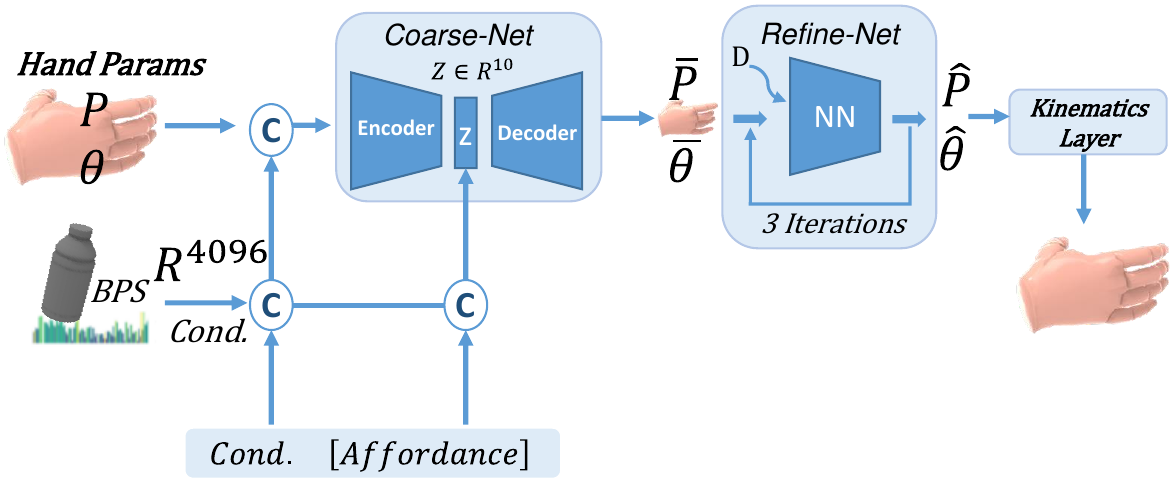}
\end{overpic}
\caption{The network architecture of AffordPoseNet for afforfance-oriented hand-object interaction generation.}
\label{fig:5_2_hand_network}
\vspace{-7pt}
\end{figure}\textbf{}
\vspace{-0.5cm}


\subsection{Affordance-oriented Interaction Generation}
\begin{table*}[ht]
\centering
\caption{The quantitative evaluations of hand-object interaction generation experiments.}
\setlength{\tabcolsep}{1.5mm}{
\begin{tabular}{c|c|ccccccccc}
\hline
\multirow{2}{*}{Metrics}                                         & \multirow{2}{*}{GrabNet} & \multicolumn{9}{c}{AffordPoseNet}                                                                                                                                                                                                                    \\ \cline{3-11} 
                                                                 &                          & \multicolumn{1}{c}{Handle-grasp} & \multicolumn{1}{c}{Lever}   & \multicolumn{1}{c}{Lift}    & \multicolumn{1}{c}{Press}   & \multicolumn{1}{c}{Pull} & \multicolumn{1}{c}{Support} & \multicolumn{1}{c}{Twist}   & \multicolumn{1}{c|}{Wrap-grasp}    & Mean     \\ \hline
\begin{tabular}[c]{@{}c@{}}Penet.Depth($cm$) $\downarrow$ \end{tabular}       & 0.87                     & \multicolumn{1}{c}{1.01}         & \multicolumn{1}{c}{0.60}    & \multicolumn{1}{c}{1.02}    & \multicolumn{1}{c}{0.09}    & \multicolumn{1}{c}{0.94} & \multicolumn{1}{c}{1.85}    & \multicolumn{1}{c}{0.97}    & \multicolumn{1}{c|}{0.94}    & 0.89    \\ \hline
\begin{tabular}[c]{@{}c@{}}Solid.Intsec.Vol($cm^{3}$) $\downarrow$ \end{tabular} & 3.20                     & \multicolumn{1}{c}{5.32}         & \multicolumn{1}{c}{2.44}    & \multicolumn{1}{c}{3.37}    & \multicolumn{1}{c}{0.97}    & \multicolumn{1}{c}{2.82} & \multicolumn{1}{c}{22.93}   & \multicolumn{1}{c}{1.83}    & \multicolumn{1}{c|}{6.04}    & 4.57    \\ \hline
Contact Ratio(\%) $\uparrow$                                                   & 96.06                  & \multicolumn{1}{c}{100}        & \multicolumn{1}{c}{100}   & \multicolumn{1}{c}{92.50} & \multicolumn{1}{c}{92.86}   & \multicolumn{1}{c}{75} & \multicolumn{1}{c}{100}   & \multicolumn{1}{c}{96.88} & \multicolumn{1}{c|}{97.26} & 96.06 \\ \hline
Affordance accuracy(\%) $\uparrow$                                                & -                        & \multicolumn{1}{c}{80}         & \multicolumn{1}{c}{72.73} & \multicolumn{1}{c}{92.50} & \multicolumn{1}{c}{95.24} & \multicolumn{1}{c}{0}  & \multicolumn{1}{c}{87.50} & \multicolumn{1}{c}{53.13} & \multicolumn{1}{c|}{98.63} & 83.51 \\ \hline
\end{tabular}}
\label{tab:metric_table}
\end{table*}

The affordance-oriented hand-object interaction generation aims to predict a possible hand pose, including the intrinsic (joint configurations) and extrinsic (hand rotation and position) parameters, from the input object and a given affordance label. Following the related work~\cite{2022OakInk}, we compare the generation ability of the original GrabNet baseline~\cite{3_GRAB:2020} and our variant named AffordPoseNet. The former is trained to predict the hand pose from the object only, while the latter takes both the object and the specified affordance condition as input. The network architecture of AffordPoseNet is shown in Figure~\ref{fig:5_2_hand_network}. We encode the affordance label as a one-hot vector and concatenate it with the object feature to generate the appropriate hand pose. Except for the 8-1-1 train-val-test split of our dataset, we add 211 objects with part-level affordance but no hand pose annotations to expand the test set.

Figure~\ref{fig:exp1} shows the qualitative evaluations of the two experiments. As expected, in the first row, although GrabNet~\cite{GRAB:2020} is trained with various hand poses for different affordances, it can only produce roughly reasonable but similar hands for each object. Taking the bottle object (the 2nd and 3rd results in the top row) as an example, the predicted hands are similar even if we sample different random vectors from the Gaussian distribution to generate the results. Actually, when one object has several affordances, the predicted hand from GrabNet~\cite{GRAB:2020} often contacts the object in the middle of the related parts, with the hand pose corresponding to the most frequent affordance, i.e. wrap grasp for the bottle cases in Figure~\ref{fig:exp1}. By contrast, the AffordPoseNet is able to predict the distinctive hand poses for the specified affordance, justifying the effectiveness of the affordance labeling in guiding hand-object interactions. 

\begin{figure}[t!]
\centering
\begin{overpic}[width=1.0\linewidth,tics=10]{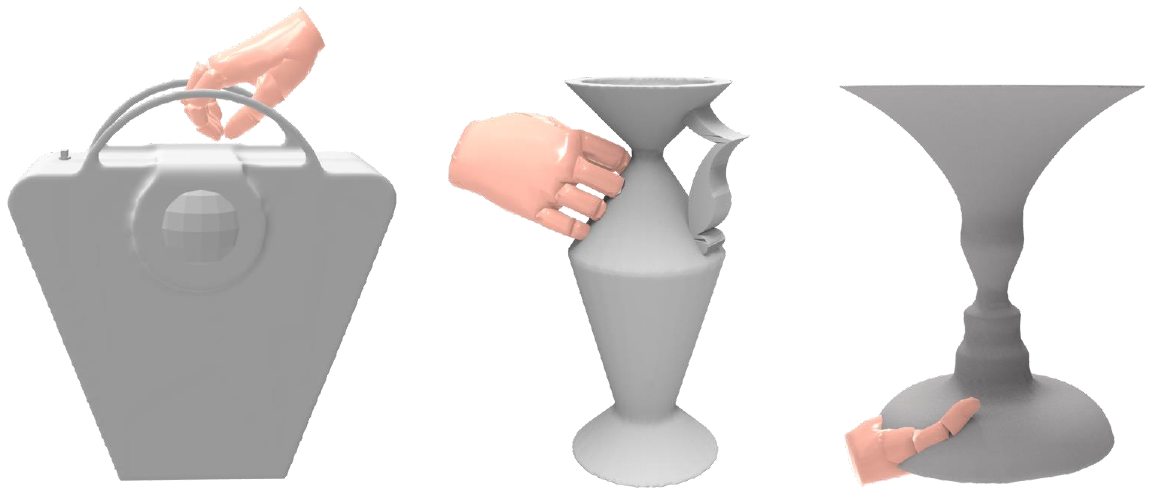}
\put(14,-3){\small Pull}
\put(46,-3){\small Handle-grasp}
\put(81,-3){\small Support}
\end{overpic}
\caption{Challenging cases produced by AffordPoseNet for the affordance-oriented interaction generation task.}
\label{fig:failure_case}
\vspace{-7pt}
\end{figure} 

Some challenging cases produced by AffordPoseNet are shown in Figure~\ref{fig:failure_case}. We can see that the generated hand poses still successfully reflect the distinctive characteristics of the corresponding affordances, in terms of the hand joint configurations and hand rotations. But the contact regions are inappropriate, resulting in unrealistic hand-object interactions. For example, for the pull affordance, the hand is usually used to interact with the zippers of bags. But it's hard for the network to predict the correct hand pose to accurately contact at this part. To solve this issue, a physical simulator could be used to post-process the generated hand, optimizing it to contact the affordance-related part.

We also adopt the commonly used quantitative metrics~\cite{11_GraspField, 2021graspTTA,12_abman} to measure the quality of the generated hand-object interactions and whether the results match the affordance conditions. The penetration depth (Penet.Depth) and solid intersection volume (Solid.Intsec.Vol) metrics reflect how much the hand models penetration with the object shape. The contact ratio is defined as the probability of the generated hands contacting the object surface. On the other hand,  the affordance accuracy metric computes whether the contacting part of the predicted hand is consistent with the ground-truth part of the input affordance condition. The contacting part is defined as the part with the most contact points ($ dis_{o2h} \leq \alpha = 0.004m $). In particular, if there's no contact point under this threshold, we gradually increase $ \alpha $ by $ step=0.001m $ until the contact points appear or $ \alpha $ reaches $0.01$. If there's no contact point when $\alpha = 0.01$, we directly mark the result as a wrongly predicted hand. Note again that we test the performance on a wider range of test set, which has the part-level affordances but without the annotated hand poses, since the above evaluation metrics don't need the ground-truth hand poses for the computation.

\begin{table*}
\centering
\caption{Quanlitative results of hand-object interaction classification.}
\setlength{\tabcolsep}{0.1mm}
\begin{tabular*}{\hsize}
{@{}p{1.8cm}|@{\extracolsep{\fill}}cccccccc|c@{}}
\hline
        & Handle-grasp  & Lever   & Lift    & Press   & Pull    & Support & Twist   & Wrap-grasp & Mean     \\ 
\hline
Precision & 97.95\% & 99.00\% & 98.97\% & 98.15\% & 96.67\% & 92.38\% & 94.53\% & 97.77\%    & 97.31\%  \\ 
\hline
Recall    & 97.63\% & 98.30\% & 99.35\% & 96.09\% & 92.06\% & 96.33\% & 96.87\% & 97.10\%    & 97.29\%  \\
\hline
\end{tabular*}
\label{tab:img_classify_result}
\end{table*}
\begin{table*}
\centering
\caption{Quanlitative results of hand mesh recovery.}
\setlength{\tabcolsep}{0.1mm}
\begin{tabular*}{\hsize}
{@{}p{1.5cm}|p{1.5cm}|@{\extracolsep{\fill}}cccccccc|c@{}}
\hline
\multicolumn{2}{c|}{}                  & Handle-grasp & Lever  & Lift   & Press  & Pull   & Support & Twist  & Wrap-grasp & Mean    \\ 
\hline
\multirow{2}{*}{\begin{tabular}[c]{@{}c@{}}Mesh\\
Recovery\end{tabular}} & MPVPE & 12.2         & 18.44  & 45.36  & 11.82  & 24.3   & 26.36   & 14.02  & 9.58       & 16.4    \\ 
\cline{2-11}
                               & MPJRE & 0.2516       & 0.2455 & 0.2278 & 0.1478 & 0.3796 & 0.1962  & 0.2043 & 0.1279     & 0.1892  \\
\hline
\end{tabular*}
\label{tab:imag-results}
\end{table*}


Table~\ref{tab:metric_table} reports the quantitative performances on the generation of hand-object interactions. Based on the top three rows, the two experiments generate the hand poses with similar quality, although AffordPoseNet has slightly worse performance w.r.t. the solid intersection volume. On the other hand, most of the results of AffordPoseNet match with the input affordance condition, while the GrabNet baseline often generates similar hand poses when the input object has multiple affordances. However, the affordance accuracies of AffordPoseNet are particularly low for the pull and twist affordances. This is because the corresponding parts, e.g. zippers of bags, lids of bottles, are relatively small in contrast with the nearby parts. When the generated hands are not accurately contacting with the appropriate region, the affordance accuracy metric often considers them as wrong predictions, although the joint configurations still show the correct distinctive characteristics as shown in Figure~\ref{fig:failure_case}. 

\subsection{Image-based Applications}

Our AffordPose dataset supports RGB-based applications as the other existing datasets do. We render the RGB images of the hand-object interactions from our dataset. Specifically, the objects are centered at the origin of the coordinate frames, and we randomly sample 3 viewpoints around each hand-object interaction instance to render the images for the RGB-based applications. 


\textbf{Hand-Object Interaction Classification.} Taking an RGB image of the hand-object interaction as input, we train a network to predict the interaction type, i.e. the affordance label. We adopt ResNet-18~\cite{Resnet} as the network architecture for this classification task. Table~\ref{tab:img_classify_result} reports the classification precision and recall to evaluate the performance. The network consistently performs well on all the hand-object interaction types. We further found a high correlation between classification performance and object functionality. The fewer affordances the object category has, the better classification it obtains. For example, the categories earphone, keyboard, knife, and laptop each have only one affordance type and gain the highest interaction classification results. For more detailed evaluation statistics, please refer to our supplementary material.


\textbf{Hand Mesh Recovery.} We also tested the hand mesh recovery task from the input RGB image, taking the part-level affordance as the input condition. To adapt our hand pose data format, the network architecture is modified from I2L-MeshNet~\cite{2020_I2L-MeshNet}. As shown in Table~\ref{tab:imag-results}, we are able to reconstruct the hand meshes from images, with MPVPE (mean per vertex position error) equal to $16.4mm$ and MPJRE (mean per joint radian error) equal to $0.1892(AUC)$ on average. Some interaction types, i.e. pull, lift, and support, have relatively worse hand pose reconstruction performance than others. This is probably due to the hand pose diversity of these interaction types, making it hard to predict the accurate hand pose with occlusion in the input image.

\section{Conclusions}

We present AffordPose, a large-scale dataset of hand-object interactions with affordance-driven hand poses. Our data analysis reveals how affordance affects the detailed configurations of the hand poses to complete the interactions. The additional affordance labeling helps to form the fine-grained hand-object interactions: the hand poses corresponding to the same affordance exhibit some distinctive characteristics as well as a certain degree of diversity. The effectiveness of our dataset is observed in the related tasks, hand-object affordance understanding and affordance-oriented interaction generation, as well as the image-based applications.

To further expand our dataset in the future, we believe that including more types of hand-object interactions with affordance labeling will stimulate a wider range of applications. For example, a series of affordance-driven dynamic interactions will demonstrate how human perform complicated tasks, such as washing face, pouring water from a bottle to a cup, etc. In addition, we are also interested in hand-hand cooperation, e.g. bi-manual manipulations, human-robot cooperation, etc, to investigate how different hands are assigned with appropriate affordances for the cooperation. In order to construct datasets with these diverse and complicated hand-object interactions, it is necessary to develop efficient hand pose annotation methods, e.g. semi-automatric algorithms, to make large-scale and high-quality data for the learning tasks.

\textbf{Acknowledgements.} We thank the anonymous reviewers
for their valuable comments. This work was supported by NSFC programs (61976040), Emerging Interdisciplinary Cultivation Project of Jiangxi Academy of Sciences (2022YXXJC0101), Shenzhen Collaborative Innovation Program (CJGJZD2021048092601003), GD Natural Science Foundation (2021B1515020085), and Shenzhen Science and Technology Program (RCYX20210609103121030). Manyi Li is also supported by the Excellent Young Scientists Fund Program (Overseas) of Shandong Province (Grant No.2023HWYQ-034).

{\small
\bibliographystyle{ieee_fullname}
\bibliography{egbib}
}

\end{document}